\documentclass{svproc}
\usepackage{times}

\usepackage[numbers]{natbib}
\usepackage{multicol}
\usepackage[bookmarks=true]{hyperref}
\hypersetup{
  colorlinks   = true, urlcolor     = blue, linkcolor    = ., citecolor   = blue }

\usepackage{geometry}
\usepackage{wrapfig}

\usepackage[acronym]{glossaries}
\usepackage{tikz}
\usepackage{verbatim}

\usepackage{algorithm}
\usepackage{algpseudocode}

\usepackage{amsthm}
\theoremstyle{plain}
\newtheorem{prop}{Proposition}
\newtheorem{define}{Definition}

\usepackage{color}

\renewcommand{\baselinestretch}{0.97}

\title{\LARGE \bf
Task and Motion Planning in Hierarchical 3D Scene Graphs
}

\titlerunning{TAMP in Hierarchical 3D Scene Graphs}

\author{Aaron Ray$^*$ \and Christopher Bradley$^*$ \and Luca Carlone \and Nicholas Roy}
\institute{$^*$ denotes equal contribution\\MIT, Cambridge, MA 02142 \\ \{aaronray, lcarlone\}@mit.edu, \{cbrad, nickroy\}@csail.mit.edu}

\begin{document}
\newcommand{\setor}{~|~}
\newcommand{\hist}{h}
\newcommand{\pa}{\texttt{PA}}

\newacronym{tamp}{TAMP}{Task and Motion Planning}
\newacronym{dof}{DOF}{degree-of-freedom}
\newacronym{pddl}{PDDL}{Planning Domain Definition Language}
\newacronym{pddlstream}{PDDLStream}{}
\newacronym{mlp}{MLP}{Multi-Layered Perceptron}
\newacronym{cnn}{CNN}{Convolutional Neural Network}
\newacronym{gnn}{GNN}{Graphical Neural Network}
\newacronym{mcts}{MCTS}{Monte-Carlo Tree-Search}
\newacronym{pwuct}{PW-UCT}{Progressive Widening for Upper Confidence Bounded Trees}
\newacronym{mdp}{MDP}{Markov Decision Process}
\newacronym{gvd}{GVD}{Generalized Voronoi Diagram}
\newacronym{rrt}{RRT}{Rapidly Exploring Random Tree}
\newacronym{cbrne}{CBRNE}{Chemical, Biological, Radiological, Nuclear, and Explosive}
\newacronym{dnf}{DNF}{Disjunctive Normal Form}

\newcommand{\chris}[1]{{\textcolor{blue}{[Chris] \emph{\bf#1 }}}}
\newcommand{\aaron}[1]{{\textcolor{green}{[Aaron] \emph{\bf#1 }}}} 

\maketitle

\begin{abstract}
Recent work in the construction of 3D scene graphs has enabled mobile robots to build large-scale metric-semantic hierarchical representations of the world. These detailed models contain information that is useful for planning, however an open question is how to derive a planning domain from a 3D scene graph that enables efficient computation of executable plans.
In this work, we present a novel approach for defining and solving Task and Motion Planning problems in large-scale environments using hierarchical 3D scene graphs. We describe a method for building sparse problem instances which enables scaling planning to large scenes, and we propose a technique for incrementally adding objects to that domain during planning time that minimizes computation on irrelevant elements of the scene graph. We evaluate our approach in two real scene graphs built from perception, including one constructed from the KITTI dataset. Furthermore, we demonstrate our approach in the real world, building our representation, planning in it, and executing those plans on a real robotic mobile manipulator.
A video supplement is available at \url{https://youtu.be/v8fkwLjBn58}.

\end{abstract}

\section{Introduction}

We aim to enable an autonomous agent to solve large-scale \gls{tamp} problems in real-world environments.
In order to do so efficiently, an abstract planning domain is needed, which accurately represents the robot's environment as well as its available actions.
Recently, significant progress has been made in the area of generating hierarchical metric-semantic representations of the world using 3D scene graphs~\cite{Hughes24ijrr-hydraFoundations,bavle2023s}.
These environmental abstractions lend themselves well to large-scale planning problems, as they are capable of storing both higher-level abstractions such as objects and connectivity of regions which are needed for task planning, as well as the low-level metric information required to check kinematic feasibility of different actions.

However, as a planning problem instance grows in the number of objects, so too does the computational burden of finding a plan. \gls{tamp} is PSPACE-Hard \cite{vega2020task}, so problems can become computationally intractable very quickly as the sizes of the state and action spaces grow \cite{tamp-survey}.
To create tractable planning problems when converting a 3D scene graph into a planning domain, it is critical to leverage the scene graph's structure and identify which elements of the environment are relevant.
Consider, for example, a robot responding to a \gls{cbrne} scenario, receiving instructions to inspect and neutralize dangerous objects scattered in a large area, represented as a scene graph. The robot can pass near an object only after it has neutralized and cleared it, and the robot may be instructed to avoid particular regions entirely.
Depending on the geometry of the scene and the specified goal, only a subset of these dangerous obstacles and regions may ultimately be relevant to finding a plan.
But, for a robot building a scene graph representation from perception, it is not at all obvious which elements of the scene graph should be added to a planning domain to ensure a valid plan can be found and executed.

Previous approaches to the problem of inferring a task-relevant planning domain have relied on representations of connectivity in the scene graph to prune superfluous elements~\cite{Agia22corl-Taskography}. However, these efforts have been limited to specific kinds of task planning problems, as the pruning approaches employed often remove information necessary for checking the geometric feasibility of plans, or they implicitly limit the types of goals that can be specified. Alternative approaches for reducing the planning problem size involve attempting to learn the relevance of planning objects, then incrementally adding objects to the domain according to the learned relevance score until the problem is solvable \cite{silver2021planning}. Unfortunately, this approach requires training on numerous similar planning problems, and is difficult to generalize to tasks at large scales in the real world.

\begin{figure*}[t]
    \centering
\includegraphics[width=0.9 \linewidth]{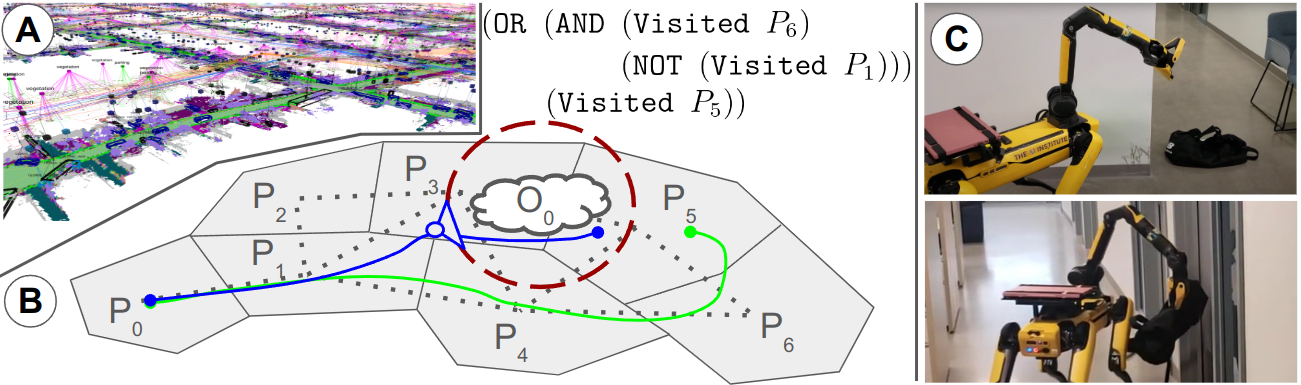}
    \caption{An illustration of how we derive and encode tasks in our planning representation from a 3D scene graph. \textbf{(A)} An isometric view of a Hydra scene graph generated from the KITTI dataset, giving an insight to the scale of the environment. \textbf{(B)} A simplified version of this scene, where the agent is tasked with either visiting Place 6 while avoiding Place 1, or visiting Place 5. We see that Place 5 is partially obstructed by a suspicious object, so the agent must consider either avoiding it (green trajectory), or inspecting and neutralizing the object (blue trajectory) to reach its goal. \textbf{(C)} A mobile robot (which we used to build our scene graphs) executing a plan in the real world, inspecting an object in the top frame, and moving an obstruction out of its path on the bottom.}
    \label{fig:front}
\end{figure*}

In this work, we propose a novel approach to both enable and accelerate \gls{tamp} in large environments (Fig. \ref{fig:front}).
Our first contribution is a three-level hierarchical planner for planning in large domains derived from 3D scene graphs.
Next, we present the formulation of a sufficient condition for removing symbols from a planning problem while maintaining feasibility, which can greatly reduce computation when planning. This condition shows that many of the places in a 3D scene graph can be ignored when formulating planning problems that factorize according to our three-level hierarchy.
We introduce a technique for reasoning about whether the sparsified domain matches the original intent of the planning domain, which reveals extra constraints that must be imposed on the motion planner.
Finally, we develop a method to further accelerate planning by incrementally identifying objects in the scene as relevant during search according to how the geometry of the scene affects the feasibility of certain high-level plans.
We show the effectiveness of our approach across two hand-crafted domains, two scene graphs built from real perception with planning in simulation, and finally a real-world mobile manipulation task on a Spot robot.

\section{Task and Motion Planning in 3D Scene Graphs}

In this section, we introduce how we encode \gls{tamp} problems, review the 3D scene graph structure that we leverage for grounding planning problems, and finally propose a CBRNE-inspired planning domain as an example of formulating a planning problem based on a 3D scene graph.

\subsection{Task and Motion Planning Preliminaries}
\label{sec:sub:prelim:task-planning}

The \gls{tamp} problem jointly considers elements of high-level task planning \cite{ghallab2016automated, karpas2020automated} and low-level motion planning \cite{lavalle2006planning} in an attempt to solve hybrid discrete/continuous, multi-modal planning problems \cite{tamp-survey}. A common formalism for encoding task planning problems is the \gls{pddl}. In a \gls{pddl} problem, a \emph{state} $\mathcal{I}$ is a set of facts, where each fact is an instance of a boolean function called a predicate $p(\bar x)\in\mathcal{P}$, which is parameterized by a tuple of symbols $\bar x = [x_1, \hdots, x_k]$ from a given set of symbols $\mathcal{O} = \{ x_i\}$. Each symbol $x_i$ is a discretization of a state variable.
Transitions between states are defined by actions $a(\bar x) \in \mathcal{A}$ (also parameterized by symbols) which are expressed as two sets of predicates: preconditions $\text{Pre}(a_i)$ and effects $\text{Eff}(a_i)$.
An action's preconditions determine if an instance can be applied from a particular state $\mathcal{I}$, and its effects define the set of facts that are added ($\text{Eff}^+(a_i))$ or removed $(\text{Eff}^-(a_i))$ from the state $\mathcal{I}$.
A planning \emph{domain} is composed of lifted sets of predicates $\mathcal{P}$ and actions $\mathcal{A}$, and a problem \emph{instance} $P = (\mathcal{P}, \mathcal{A}, \mathcal{O}, \mathcal{I}_0, \mathcal{G})$ combines a domain with an initial state $\mathcal{I}_0$ and a set of goal states $\mathcal{G}$, parameterized by symbols $\mathcal{O}$.

Solutions to \gls{pddl} problems take the form of a sequence of parameterized action instances $\pi = [a_1(\bar x_1), a_2(\bar x_2), ..., a_n(\bar x_n)]$ where the state after taking each action satisfies the precondition of the following action~\cite{tamp-survey}.
For any action sequence, there is a corresponding sequence of states $\mathcal{I}_\pi = [\mathcal{I}_0, \mathcal{I}_1, \mathcal{I}_2, ..., \mathcal{I}_n]$, leading from the initial state to a goal state that can be constructed from each action's effects.
We will use the fact that only a subset of state symbols are needed for each action to enable a factorization of the planning problem in Section~\ref{sec:hierarchical_planning}.
For an action plan $\pi$, its corresponding state plan $\mathcal{I}_\pi$ is \emph{valid} if $\mathcal{I}_i \in \text{Pre}(a_{i+1})$ for $i=0,...,N-1$, and $\mathcal{I}_N \in \mathcal{G}$.
A range of solvers \cite{fast_downward,fast_forward} can solve tasks specified in PDDL, and any state plan found by such a solver is valid by construction. A feasible planning problem is one for which there is a valid solution.

The continuous nature of \gls{tamp} problems, coupled with the scale of environments we consider here, make discretizing and encoding a planning problem directly in pure \gls{pddl} infeasible. We therefore use an extension of \gls{pddl} called PDDLStream~\cite{pddlstream} to represent and solve \gls{tamp} problems.
A PDDLStream problem instance $(\mathcal{P}, \mathcal{A}, \mathcal{S}, \mathcal{O}, \mathcal{I}_0, \mathcal{G})$ represents the discrete search portion of a \gls{tamp} problem in \gls{pddl}, as a set of predicates, actions, symbols, initial state, and set of goal states, but also introduces the notion of \emph{streams} $s \in \mathcal{S}$, which can query external samplers/solvers (e.g. a motion planner) during search to produce new symbols and facts within the problem instance. Streams make the problem encoding more efficient, as they obviate the need to evaluate predicates for all possible continuous values of a symbol. PDDLStream solves\footnote{Specifically, the PDDLStream \emph{adaptive} solution algorithm.} problems by first finding an optimistic solution that satisfies the domain's symbolic constraints -- a \emph{task skeleton} -- and then attempting to solve for feasible continuous parameters. We refer the reader to \cite{pddlstream} for a detailed description of PDDLstream.

In a \gls{tamp} problem,
each symbol $x_i$ can represent a continuous value (e.g., a pose), and the grounded parameters of an action depend on these values. From these parameters, we can derive a \emph{motion sequence}, which specifies how a robot executes an action. For example, from an action plan composed of a sequence of \texttt{move} actions, the corresponding motion sequence would be composed of the trajectories that were solved for by the motion planner and describes the continuous values of the parameters $\bar{x}_i$ for each \texttt{move} action. Executing that sequence involves multiple calls to a trajectory controller, where two motion sequences are equivalent if they result in the agent acting identically.

\subsection{Building 3D Scene Graphs from Perception}

A robot operating in the real world ideally is able to build its own PDDL domain model from perception. We assume the robot is equipped with a prior model of its own state and a motion controller, as well as the ability to use its sensors to build a dense geometric model of its environment and the objects in it.
To derive a discrete, symbolic model of the geometry, we take advantage of recent work in 3D scene graph mapping~\cite{Hughes24ijrr-hydraFoundations} that infers a discretization of the geometry and the objects in the geometric model.
While our approach is compatible with a range of scene graph implementations, our definition of a 3D scene graph, directly based on Hydra ~\cite{Hughes24ijrr-hydraFoundations,strader2023indoor}, consists of several layers of increasing abstraction (see Fig. \ref{fig:front}). Each layer consists of a collection of nodes representing location and other attributes, with edges connecting nodes within the same layer representing relative spatial constraints and edges between different layers representing an inclusion relationship. The lowest layer of the hierarchy is a semantically-annotated \emph{mesh} of the scene geometry. The next layer contains \emph{objects} and their locations identified by a semantic image segmentation. The \emph{places} layer represents navigable regions of the environment based on semantic and geometric properties of the mesh. Places are clustered into groups based on geometric and semantic information, and these groups become nodes in the higher-level \emph{regions} layer (e.g., rooms in an indoor environment). Hydra can construct this map representation in realtime from RGBD sensor data while accounting for odometry drift, enabling large scale, consistent, and information-rich maps.

Previous work on 3D scene graphs has mainly focused on indoor uses. These representations rely on the \gls{gvd}~\cite{Oleynikova18iros-topoMap} to generate places, an abstraction of 3D spatial connectivity, which are not well suited for ground robot navigation. We use an alternate formulation of 2D places in our navigable scene graph, where each place represents a 2D polygon with consistent terrain classification, representing an area the robot may traverse (Fig.~\ref{fig:front}B).

\subsection{Inferring the Planning Domain from Scene Graphs}
\label{sec:sub:prelim:direct}

We introduce a framework for deriving a \gls{tamp} problem instance from a scene graph to demonstrate the salient aspects of solving planning problems based on large-scale environments. In general, the problem contains a symbol $x \in \mathcal{O}$  for each node in the scene graph, as well as a symbol corresponding to the robot. We define six classes of predicates, derivable from a Hydra scene graph, that may be relevant for planning:
1) Type information derived from the nodes of the graph, where each node corresponds to a unary predicate: \texttt{(Configuration ?c)}, \texttt{(Place ?p)}, and \texttt{(Object ?o)}, etc.
2) Agent or object predicates that define the state of the robot and objects: \texttt{(AtConfig ?c)}, \texttt{(AtPlace ?p)}, \texttt{(AtRoom ?r)}, etc.
3) Connection predicates defined by edges in the same level of the graph: \texttt{(Connected ?n1 ?n2)},
4) Inclusion predicates indicating edges connecting nodes of different levels of the graph: \texttt{(PoseInPlace ?c ?p)}, \texttt{(PlaceInRoom ?p ?r)}, etc.
5) Preconditions of actions that are certified by solving a stream's associated sub-problem. For example, a \textit{move} action
may require that a trajectory has been found between two configurations: \texttt{(Trajectory ?c1 ?t ?c2)}.
6) Additional, problem-specific predicates defined by the user to specify goal states and problem constraints such as which places to visit or which objects to collect.

As a running example, we define an example problem using these predicates, motivated by \gls{cbrne} scenarios. In our ``Inspection Domain", an agent can be commanded to visit or avoid certain places, and inspect and neutralize objects that have been marked as suspicious. The robot cannot move past a suspicious object until it has been inspected and neutralized. We therefore define problem specific predicates: \texttt{(VisitedPlace ?p)} which indicates the current and past places the robot has been to, and \texttt{(Safe ?o)} or \texttt{(Suspicious ?o)} which describe an object. We define streams for sampling poses for inspecting and neutralizing objects, sampling poses in a specific place, and planning motion between two poses in order to find feasible continuous parameters for a given abstract plan.
Goal specifications in this domain can include positive or negated facts based on these predicates. The agent's available actions are to \texttt{move} between poses in connected places, and to \texttt{inspect} objects from appropriate poses (for simplicity we do not separate the inspect and neutralize actions). Note that only \texttt{move} needs to be parameterized by a place symbol. Any valid plan is composed of these actions, which at execution time are converted into a motion sequence of \texttt{FollowPath}($t_i$) and \texttt{InspectObject}($o_i$) primitives for paths $t_i$ and objects $o_i$.

\section{Scalable Scene Graph Planning}

Our objective is to both enable and accelerate solving \gls{tamp} problems in large environments. One associated challenge is that the number of symbols created by a scene graph can quickly overwhelm the ability of the planner to reason efficiently due to increasing branching factor, and the depth of search needed to find plans.
However, we notice that the vast majority of planning domains, and the world in general, tend to factor into sequence of navigation actions punctuated by periodic object-centric actions.
This factorization allows us to identify a subset of symbols relevant for object interaction, and a subset needed to move from place to place, potentially simplifying search.
We therefore propose a planning formulation that aligns with the scene graph hierarchy and naturally divides the planning problem into a high-level, task-relevant planning problem such as finding a sequence of manipulation actions, mid-level coarse navigation planning between locations, and low-level continuous trajectory planning between points.

Specifically, instead of requiring a PDDL planner to find paths through the places in the scene graph at the discrete symbolic level, we reduce the depth of the planning horizon at the highest level by reasoning only over places that are directly relevant for achieving the goal. Then, a coarse navigation planner plans through the 2D places layer to create an abstract motion plan composed of a sequence of subgoals. Finally, a fine-grained motion planner is guided by the navigation plan subgoals through the places layer, quickly finding motion plans over large distances. We discuss this in Sec.~\ref{sec:hierarchical_planning}.

By focusing each layer of our tri-level planner hierarchy on specific types of actions, we can prune irrelevant symbols within each layer and simplify the corresponding problems by reducing the branching factor of search (Sec.~\ref{sec:prune}).
Critical to our approach is that the proposed factorization must not limit the types of problems that can be solved, nor produce plans which violate intended constraints. To that end, we show a sufficient condition for symbols to be removed from each planning problem while maintaining feasibility, then show how to reason about whether the resulting motion plans adhere to the original specification (Sec.~\ref{sec:execution-consistency}).
Finally, in Sec~\ref{sec:object_pruning}, we consider an additional heuristic that enables optimistically ignoring objects that are irrelevant due to scene geometry.

\subsection{Hierarchical Planning}
\label{sec:hierarchical_planning}

\begin{figure*}[t]
    \centering
    \includegraphics[width=0.99\linewidth]{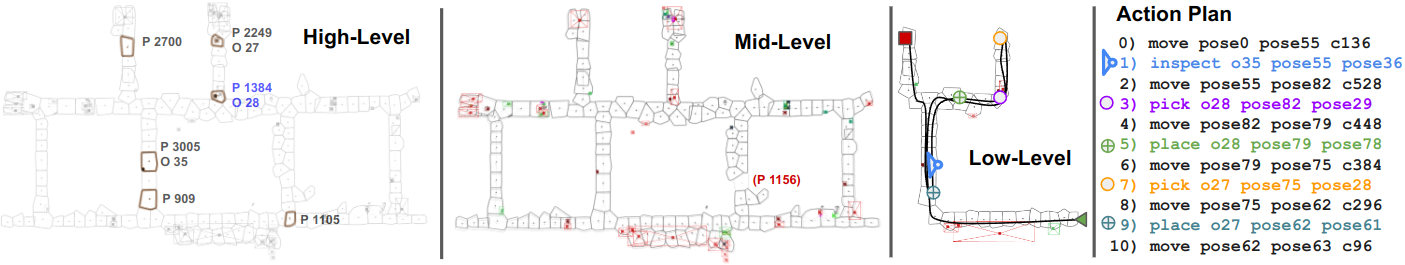}
    \caption{Our tri-level planner in a real world environment. A scene graph of the fifth floor of building 45 at MIT was built using the Spot robot, from which we extract our planning abstraction for the goal: \texttt{(and (ObjectAtPlace O27 P909) (VisitedPlace P2700) (Safe O35) (not (VisitedPlace P1153)))}, which instructs the agent to move \texttt{O27} to a \texttt{P909}, inspect \texttt{O35}, visit \texttt{P2700}, all while avoiding \texttt{P1153}. At the highest level, the task planner is given a very sparsified version of the scene, as highlighted above. The mid-level planner plans a path through the places guided by the abstract plan found at the highest level, avoiding place \texttt{P1153}.  Feedback from this level leads to the addition of \texttt{O28} to the high-level domain, as \texttt{O27} would be otherwise unreachable. The low-level planner computes full trajectories, guided by the path found at the mid-level. The plan produced (and executed by the robot) is shown on the right.}
    \label{fig:manipulation-experiment}

\end{figure*}

Here we describe our tri-level planning approach. The highest-level of the planner must reason about which objects to interact with, which regions to avoid, and which destinations the robot should move to.
The most straightforward encoding of planning motion through a scene graph would closely mirror the connectivity of the graph, modeling the \texttt{move} action as a transition between two places that share an edge in the graph. We refer to this problem encoding as the \emph{direct encoding} of the domain. However, as the scale of the environment increases, this direct encoding results in very long horizon plans that are expensive for the general-purpose PDDL planner to find.

Instead, we propose a more general move action: \texttt{moveRelaxed}.
This action takes place $p_1$ and $p_2$ as parameters, in addition to
initial and final poses $c_1$ and $c_2$ and a trajectory symbol $t$. The
action's effect moves the robot's pose from $c_1$ to $c_2$, and marks $p_1$ and $p_2$ as visited.
For example, consider the task in Fig.~\ref{fig:front}B where the robot begins in Place $P_0$, and is tasked with visiting place $P_5$. A motion sequence corresponding to a plan to move from place $P_0$ to place $P_1$, then from $P_1$ to $P_4$, and finally from $P_4$ to $P_5$ can be equivalent to a sequence generated from a plan to move from $P_0$ to $P_5$ directly, meaning the high-level planner need not explicitly plan to move through $P_1$ and $P_4$. Extrapolating this pruning approach to larger scenes and more complex goals has the potential to vastly reduce planning horizons, although it imposes certain constraints on the lower-level planners that will be addressed at length in Section~\ref{sec:execution-consistency}.

Abstract plans produced by the high-level planner do not initially contain information about how the robot moves from the start to end poses.  Instead, they optimistically contain trajectory symbols with continuous parameters that must be filled in by the lower-level planners. In order to do this efficiently, we rely on the 3D scene graph to accelerate motion planning.
To find a motion plan between two configurations $c_1$ and $c_2$, we plan through the places layer of the scene graph, finding a sequence of places that leads from $c_1$ to $c_2$ and respects the connectivity of the scene graph. This level of abstraction can take advantage of Euclidean distance heuristics to accelerate planning, while allowing for the constraints of the task (e.g., avoiding a particular place) to be encoded simply.

At the lowest level of abstraction, the planner generates a kinematically-feasible path for the robot to follow based on the reference path from the mid-level planner. This path can be generated efficiently by first considering an optimistic path that connects the waypoints on the reference path and ignores obstacles. Any segments of this path that are rendered infeasible by obstacles or violations of kinematic constraints can be re-solved by a planner that considers obstacles, such as RRT~\cite{lavalle1998rapidly}.
Better alignment between which edges are present in the scene graph and kinematic feasibility for the robot leads to better performance of this heuristic. Fig.~\ref{fig:manipulation-experiment} highlights our tri-level planner in the real world.

\subsection{Removing Redundant Symbols}
\label{sec:prune}

Our relaxed encoding reduces the planning depth significantly for the high-level planner, but it introduces a different problem of branching factor.  Now the high-level planner can consider moving between any two places as a valid action, which can make search difficult in certain problem instances. To address this concern, we can reduce the size of the planning instance by pruning \texttt{Place} symbols that are not relevant to a given planning problem. However, identifying symbols that do not impact the solution is in general as hard as solving the problem itself, and naively removing places from a problem instance might render the problem infeasible. In this section, we characterize a set of places that we know can safely be removed from the problem before planning given the semantics of \texttt{moveRelaxed}.
We begin by defining a set of symbols that are redundant for a particular goal specification.
\begin{define}[Redundant Symbol]\label{def:strong_redundant}

    For a set of domain actions $\mathcal{A}$ and specific goal $\mathcal{G}$, a symbol x is redundant if both of the following hold:
    \begin{enumerate}
\item For every valid plan $\pi$ where x parameterizes an action, there is another valid plan $\pi'$ with equivalent motion sequence, where x is not an action parameter,
        \item No action precondition or goal, expressed in negative normal form, contains a universal quantifier that can be parameterized by x.
    \end{enumerate}

\end{define}
\noindent The intuition behind this notion of redundancy is that
1) if any plan involving the symbol yields a motion sequence that can be rewritten without the symbol, the symbol is redundant, and 2) if we solve a planning instance where a redundant symbol has been removed, we would like to know that the plan is still valid in the original problem.
Note that this definition of redundancy is general for any planning domain, although we will use this definition specifically for place symbols that become redundant given the \texttt{moveRelaxed} action. Importantly, removing redundant symbols preserves the feasibility of a planning problem.

\begin{prop}[Removing Redundant Symbol Preserves Feasibility]\label{prop:ignore_strongly_redundant}

Consider a feasible planning instance $R = (\mathcal{P}, \mathcal{A}, \mathcal{S}, \mathcal{O}, \mathcal{I}_0, \mathcal{G})$. For a redundant symbol $x \in \mathcal{O}$, we define a related instance $R' = (\mathcal{P}, \mathcal{A}, \mathcal{S}, \mathcal{O}', \mathcal{I}_0', \mathcal{G}')$ where $x$ has been removed, i.e., $\mathcal{O}' = \mathcal{O}\setminus x$ and $\mathcal{I}_0'$ contains all facts in $\mathcal{I}_0$ except those parameterized by $x$, and similarly for $\mathcal{G}'$. Let $\Pi_R$ denote the set of valid plans for $R$. Then, $\Pi_{R'} \subseteq \Pi_R$ and $\Pi_{R'}\neq \emptyset$. (Proof deferred to the appendix.)

\end{prop}

\noindent The requirements for a symbol to be redundant are quite strong (\emph{every} plan that uses a symbol must have an alternate plan that does not use the symbol and  still results in the same motion sequence), but many places in the Inspection Domain have this property given the semantics of \texttt{moveRelaxed}.
Removing these places from our task planner's domain, assuming the motion planner is still aware of them, enables the solver to more efficiently find valid plans that are guaranteed to have also been valid in the un-pruned problem. Moreover, the ability to prune these elements does not restrict the type of goals we are able to specify to our agent, preserving expressivity, while enabling planning at a larger scale.

\begin{prop}[Redundant Places]\label{prop:redundant_places}
Consider a problem instance in the Inspection Domain with no quantifiers that can be parameterized by a place in the goal. A place $p$ is redundant if no facts parameterized by $p$ appear in the initial or goal states, or if \texttt{(not (VisitedPlace $p$))} appears as a clause in the conjunctive normal form (CNF) of the goal specification. (Proof deferred to appendix.)
\end{prop}

\noindent We have now identified a potentially large (depending on the sparsity of the goal specification) set of place symbols that can be ignored in the Inspection domain. Our explicit method of defining our problem's initial state is as follows:

\begin{remark}[Problem Initialization]\label{rem:place_initialization}
In light of Proposition \ref{prop:redundant_places}, we only include the following places when instantiating a problem in the Inspection domain: 1) the initial place that the robot is in and 2) any place that appears in the goal. A place $p$ that parameterizes a negated fact $(\texttt{not } (\texttt{VisitedPlace } p))$ that appears as a clause in the CNF of the goal specification can also be removed.

\end{remark}

\noindent We have shown that for the Inspection domain, redundant places are very easy to identify and that excluding them from a planning instance adds no computational overhead at runtime. We would like to apply the same idea to similar domains, without needing to reason from scratch about redundancy. We now characterize a sufficient condition on the planning domain structure for places to be redundant. Let $\mathcal{P}_{static}$ denote the set of predicates that do not appear as effects of any action (i.e., they can only be set in the initial state). Let $\mathcal{F}_{static}$ denote the set of facts that correspond to parameterizations of $\mathcal{P}_{static}$. Intuitively, if a domain is structured such that a place can only be parameterized by an action if certain facts hold in the initial state, then it is very easy to check whether a specific place can be used by any actions.

\begin{prop}[Sufficient Conditions for Ignoring Places]

Consider a planning instance $(\mathcal{P}, \mathcal{A}, \mathcal{S}, \mathcal{O}, \mathcal{I}_0, \mathcal{G})$, where for all actions $a_j \in \mathcal{A}$ except $a_j = \texttt{moveRelaxed}$, satisfying $\text{Pre}(a_j)$ implies that any place parameterized by $a_j$ is in $F_{static}$. In this case, all places that do not parameterize any facts in $\mathcal{F}_{static}$ or the goal are redundant.

\end{prop}

\noindent For example, if the Inspection domain is augmented with a ``report home" action that can only be executed at a designated set of places, then these places (and no others) need to be added to the problem instance. Fig.~\ref{fig:manipulation-experiment} illustrates how we prune the planning domain.

\subsection{Execution Consistency}
\label{sec:execution-consistency}

While our decision to use \texttt{moveRelaxed} to model motion between distance places enables faster planning, it creates a mismatch between the logical and continuous parts of the problem. In the example in Fig.~\ref{fig:front}B, consider that the robot was also instructed to avoid place $P_1$. The ability to include a constraint on the goal states of the form \texttt{(not (VisitedPlace $P_1$))} requires further constraints on the mid- and low-level planners. Executing \texttt{moveRelaxed} from place $P_0$ to place $P_6$ may involve following a trajectory that takes the robot through place $P_1$, even if the goal specifies that $P_1$ should not be visited. Technically this is still a valid solution to the planning problem since place $P_1$ never appears as a parameter to the \texttt{moveRelaxed} action (and therefore \texttt{(VisitedPlace B)} is not an effect), but clearly the domain with a relaxed movement action does not fully capture the intent of the original planning domain.

To formalize the discrepancy between what happens when the robot executes a motion sequence and the constraints that we expect a planning problem to impose, we introduce the concept of a \emph{verifier function}. A verifier function maps motion sub-sequences to sets of \gls{pddl} domain facts, and ``verifies'' which additional domain facts would be implicitly true as a result of the agent executing a motion sequence, even if actually adding these facts to the problem instance during the solving process is undesirable computationally.
Given a verifier $V$, the facts that hold at each step when executing a motion sequence may be different than expected in the original plan. We denote the facts that would be added by such a verifier applied to the motion sequence associated with $a_i$ as $V(a_i)$, and term this sequence of expanded states the $V$-extended state plan.

\begin{define}[V-Extended State Plan]
For an action plan $\pi = [a_1, ..., a_n]$, its corresponding state plan $\mathcal{I}_\pi = [\mathcal{I}_0, ..., \mathcal{I}_n]$, and verifier function $V$, the V-extended state plan for state $\mathcal{I}_k$ is $\mathcal{I}'_1 = \mathcal{I}_1 \cup V(a_1)$, and $\mathcal{I}'_k = \mathcal{I}_k \cup \left(\mathcal{I}'_{k-1} \setminus \text{Eff}^-(a_k)\right) \cup V(a_k). $

\end{define}
\noindent The extended state $\mathcal{I}'_k$ is the state at step $k$ as experienced by the verifier. $\mathcal{I}'_k$ is composed of the facts $\mathcal{I}_k$ in the initial plan, plus any extra facts that were present in the previous extended state $\mathcal{I}'_{k-1}$ other than those removed by action $a_k$, plus any facts that would be returned by a verifier applied to action $a_k$. As discussed in Sec.~\ref{sec:sub:prelim:task-planning}, any state plan found by a search algorithm is valid by construction.  However, a state plan that is augmented with the extra facts that would be produced by a verifier might not be valid.

Consider a verifier $V_{place}$ that takes a motion sub-sequence $\mu$, and returns a VisitedPlace fact for each place that intersects with the agent's position while executing $\mu$. For a place $p$ and a trajectory $t$ to be followed by the motion primitive $\texttt{FollowPath}(t)$, we denote $p \cap t$ the section of $t$ that intersects with $p$. We can then define a verifier as
\begin{align}
V_{place}(\mu) = \{&\texttt{(VisitedPlace p)}~|~
&p \cap t_i \neq \emptyset \text{ for } \texttt{FollowPath}(t_i) \in \mu  \}.
\end{align}
If the motion sequence associated with the action plan would result in the agent visiting a place that we do not expect, then the $V_{place}$-extended state plan would include a VisitedPlace fact that may conflict with the goal. If we care about the robot's motion respecting the problem's constraints on visiting certain places, then we need to prove that the $V_{place}$-extended state plan is a valid solution to the planning problem for any instance of the planning domain.

From this idea, we define the concept of execution consistency, which requires that solutions to the planning problem are still valid after considering the facts from a verifier.
\begin{define}[Execution consistent]
A domain is execution consistent with respect to verifier $V$ if, for every valid plan $\pi$, the V-extended state plan is valid.
\end{define}
\noindent A domain is trivially execution consistent for the empty verifier $V(\cdot) = \emptyset$, as the extended state plan is equal to the original plan. A domain is also execution consistent if the range of $V$ applied to each motion subsequence $\mu_i$ corresponding to action $a_i$ is limited to facts in $\text{Eff}(a_i)$. In other cases, a domain can still be execution consistent for a verifier that would introduce new facts if the domain is carefully crafted. In defining a planning domain for any task, we seek to have it execution consistent with respect to any defined verifiers. If a domain is not execution consistent, then any properties related to predicates in the verifier cannot be guaranteed to hold when executing a plan.

In our example, we want to prevent the agent from entering places that it should not, and so we should show that the Inspection domain is execution consistent with respect to the verifier $V_{place}$. Recall that $V_{place}$ can only introduce new \texttt{VisitedPlace} facts. As \texttt{VisitedPlace} does not appear in any action preconditions, the only way for a \texttt{VisitedPlace} fact to render a valid state plan invalid is to conflict with the goal specification. Consider the set of places that must be avoided to satisfy some goal state: $\mathcal{P}_{avoid} = \{p ~|~ \texttt{(not (VisitedPlace p))} \in \mathcal{G}\}$. If a place in $\mathcal{P}_{avoid}$ can only be visited by an action that explicitly lists it in the action effects, then the domain will be execution consistent with respect to $V_{place}$. This can easily be guaranteed by preventing the motion planner from generating plans that enter places $\mathcal{P}_{avoid} \setminus \mathcal{P}_{param}$, where $\mathcal{P}_{param}$ is the set of places that appear as parameters to the action.  Our mid- and low-level motion planners are constrained not to enter a place which we might want to avoid, unless that place is given as the parameter to the $\texttt{moveRelaxed}$ action, ensuring execution consistency. Note that the verifier need not be actually implemented, but the concept can be used to prove execution consistency.

\subsection{Ignoring Irrelevant Objects}
\label{sec:object_pruning}

We have demonstrated the ability to identify and ignore elements of a planning instance that are redundant when searching for a plan. However, many symbols are not redundant according to our definition, but might still be safely ignored. Objects which might obstruct motion, for example, are not redundant because an ``inspect" motion primitive will never be generated for an object that has been removed from the planning instance. Nevertheless, there are clearly cases when an agent can ignore objects when planning, such as an object that is not part of an agent's goal and is far from the agent's path to the goal. As with the places, ignoring objects can accelerate planning by reducing the branching factor at the highest level of search. In contrast to places however, the objects that should be ignored cannot be identified from the logical structure of the planning instance alone; the problem geometry must also be considered.

We propose an incremental approach to identifying relevant symbols. We begin by including some subset of all symbols $\mathcal{O}_S$ in the domain of the high-level planner, and attempt to solve the planning problem. If this limited problem has a valid solution which is also a valid solution to the original problem, then we have found a plan. Otherwise, we incrementally add symbols to the planning problem, and repeat (Algorithm~\ref{alg:incrementalObjectSolver}). The inner loop (Line 7) corresponds to solving a \gls{tamp} instance with symbols $\mathcal{O}_I$.
The outer loop (Line 5) corresponds to adding more symbols to $\mathcal{O}_I$ when we fail to find a solution.

The performance of this incremental planning approach depends on three key choices: which initial symbols are chosen in $\mathcal{O}_S$, when new symbols are added to the planning instance, and how the new symbols $\mathcal{O}_{new}$ are chosen. As long as all symbols are eventually added to the planning instance, this planning approach will maintain the completeness properties guaranteed by the chosen PDDLStream solution algorithm~\cite{pddlstream}.

\begin{algorithm}[H]
\caption{Incremental Object Solver}\label{alg:incrementalObjectSolver}
\begin{algorithmic}[1]
\Procedure{IncObjSolver}{$A, S, \mathcal{O}, I ,G$}
\State $\mathcal{O}_{S} \leftarrow GetRelevantObjects(\mathcal{O})$
\State $\mathcal{O}_{I} \leftarrow \mathcal{O}_S$
\State $\mathcal{O}_{R} \leftarrow \mathcal{O}\setminus \mathcal{O}_S$
\While{$|\mathcal{O}_I| < |\mathcal{O}|$}
\State $SkelInfo = [~]$
\For{$k \in Skeletons(A, S, \mathcal{O}_I, G)$}
    \State $T \leftarrow SolveSubProblems(k)$
    \State $Feedback \leftarrow Check(T, \mathcal{O}_R) $
    \State $SkelInfo.\texttt{append}(Feedback)$
\EndFor
\If{$\pi \in SkelInfo$ is valid}
 \State \textbf{return} $\pi$
\Else{}
\State $\mathcal{O}_{new} = NewObj(SkelInfo)$
\State $\mathcal{O}_I = \mathcal{O}_I \cup \mathcal{O}_{new}$
\State $\mathcal{O}_R = \mathcal{O}_R \setminus \mathcal{O}_{new}$
\State
\EndIf
\EndWhile
\State \textbf{return} INFEASIBLE
\EndProcedure
\end{algorithmic}
\end{algorithm}

The initial set $\mathcal{O}_S$ should be as small as possible while still including the symbols necessary to find a plan. In particular, we can begin by including symbols based on the problem's logical structure. For the Inspection domain, we include the non-redundant places identified in Remark~\ref{rem:place_initialization} and any objects that appear in the goal. In general, there is a large body of literature dedicated to identifying object relevance, such as by reachability analysis~\cite{blum1997fast, fishman2023task} or learning to predict importance~\cite{silver2021planning}, which may identify more symbols to add to the initial set.

We must decide how many task plan skeletons will be checked by $Skeletons$ and how much time will be spent attempting to solve the continuous subproblems before adding new symbols to the planning instance. In the Inspection domain, the full problem only has a solution if the pruned problem has a solution, so we choose to stop iterating through plan skeletons once we find a solution to the reduced problem. In general, a maximum time must be set for iterating through plan skeletons (we do so according to the Adaptive approach \cite{pddlstream}).
Finally, the choice of symbols to add to the planning instance can be informed by feedback ($Check$, Line 9) from failed solutions to subproblems. In our domain, we add objects which block the robot's motion on an otherwise-feasible trajectory.

\section{Evaluation}
\label{sec:eval}

\begin{figure*}[t]
\centering
    \includegraphics[width=0.9\columnwidth]{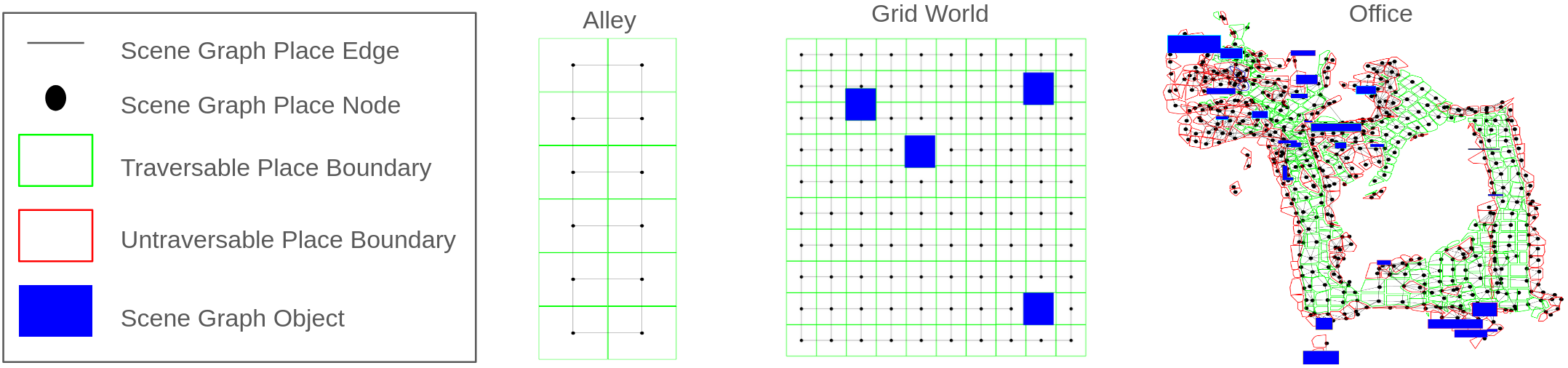}
    \caption{Three of the maps used for evaluation (not shown is the KITTI or building 45 environments). A narrow alley map, a simple 10x10 grid world map, and a scene graph built from real data collected by a robot in an office environment.}
    \label{fig:map-placeholder}
\end{figure*}

We characterize how our method's performance depends on goal complexity, environment scale, and scene geometry.
We compare our encoding of the Inspection domain to the dense, direct encoding in a variety of different settings. We test on four map archetypes (Fig. \ref{fig:map-placeholder}) -- a synthetic small constrained alleyway, a synthetic 10x10 gridworld, a scene graph built from real data in an office environment comprising 557 Places and 28 Objects, and a much larger scene graph built from the KITTI dataset composed of 17861 Places and 1315 Objects (Figs.~\ref{fig:front} and \ref{fig:map-placeholder}). For each environment, we report planning times for several different goal clauses across different variations in robot and object initial conditions. The planning time includes the incremental detection of relevant objects and the PDDL solver's preprocessing of the planning problem. The time to convert a scene graph into the planning problem is insignificant compared to the planning time. Finally, we also implement and test our planner on a Spot robot.

To randomize tasks across trials, we define a mechanism for sampling goal specifications according to an increasing number of clauses in \gls{dnf}.  A goal with complexity (N, K) is a formula in \gls{dnf} with N clauses, where each clause has K conjunctions. For example, for complexity (2, 3), the goal has the form \texttt{(Or ($C_1$, $C_2$))}, where $C_i$ is a clause consisting of three facts e.g., \texttt{(And ((Visited P1), (Safe O4), (Not (Visited P9))))}.

\begin{figure*}[t]
    \centering
\includegraphics[width=0.9\columnwidth]{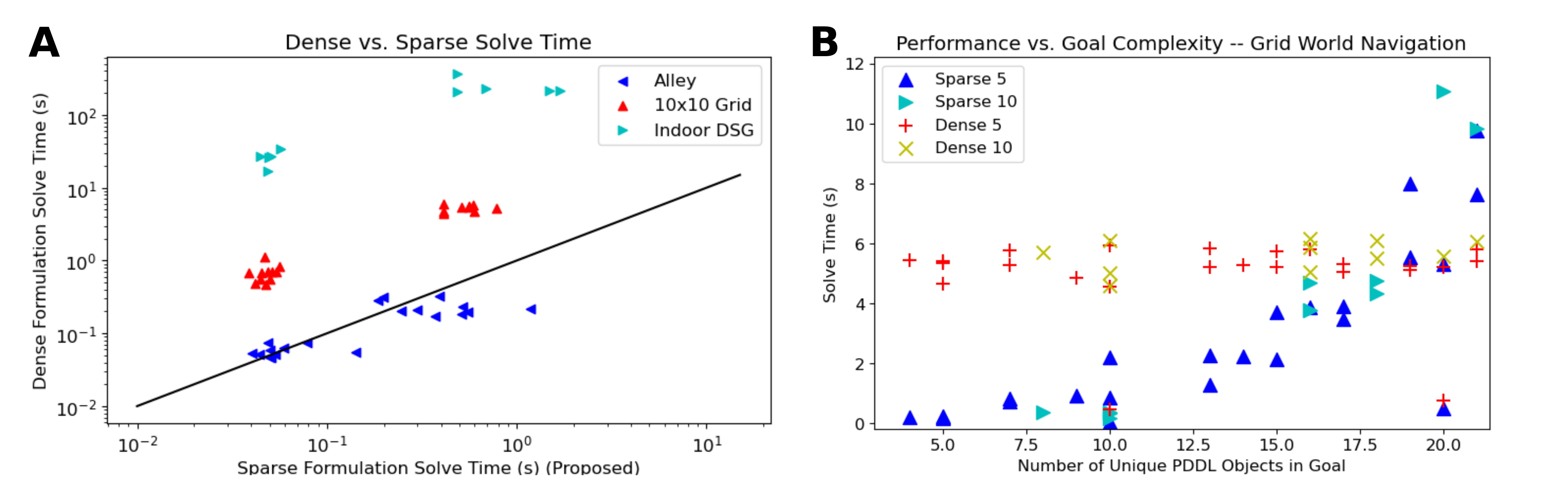}
    \caption{\textbf{A)} Comparison of the time to solve tasks of comparable complexity across different environmental scales for the dense formulation and the proposed sparse formulation. \textbf{B)} The scaling of our approach with the complexity of the goal specification in the simple 10x10 grid world. As we increase the number of unique \gls{pddl} objects in the goal specification, the problem is no longer sparse, and so it no longer benefits from our approach.
    }
    \label{fig:scaling}
\end{figure*}

\noindent\textbf{Scene Graph Size:} First, we investigate the effect of scene graph size on our ability to plan. For this set of trials, the goal complexity is (N, K) = (3, 3), and we compare planning time for the direct encoding to the planning time for our planner, shown in Fig.~\ref{fig:scaling}A. Each point on the scatter plot corresponds to a single trial and different colors correspond to different environment types. Any samples above the black line indicate that our planner outperforms the dense baseline.
In the small Alley environment, our planner performs about as well as the dense encoding, as there is not much advantage to sparsification in such a small environment.  As we scale up however, the relative performance of our planner improves. In the 10x10 grid, we see modest improvement as shown by the red points in Fig.~\ref{fig:scaling}A.  As we scale up even further, with the small-scale scene graph, we see the baseline planner taking hundreds of seconds to plan, while our planner averages in the tens of seconds. This experiment only considers goals that involve visiting or not visiting certain places in the scene graph. When we attempted to introduce object inspection, the dense baseline planner timed out before finding a plan in almost all instances. Similarly, when testing the baseline in the KITTI scene graphs, it was also unable to find solutions for goals of any complexity. Our proposed planner experienced only a modest increase in planning time as the size of the map scaled.

\noindent\textbf{Goal Complexity:} Next, we consider the effect of increasing goal complexity on planning time. To do this, we investigate a series of different goal constructions in the Grid World environment. Specifically, we run experiments with goal complexity (N, K), for K = 5 and K = 10, and N from 1 to 5. Goal facts are chosen to be either visiting or not visiting specific places.
Fig. \ref{fig:scaling}B presents a plot comparing the complexity of the goal in terms of total unique symbol referenced vs planning time. For less complex goals in this environment, our planner outperforms the dense planner, up to a crossover point at around 20 unique objects. Advantages from the additional structure in the dense formulation outweigh the gains of our sparser method when a large percentage of the place symbols are relevant to the goal.
The direct encoding never successfully completes a trial in the KITTI dataset due to timing out.

\begin{figure*}[t]
    \centering
\includegraphics[width=.6\columnwidth]{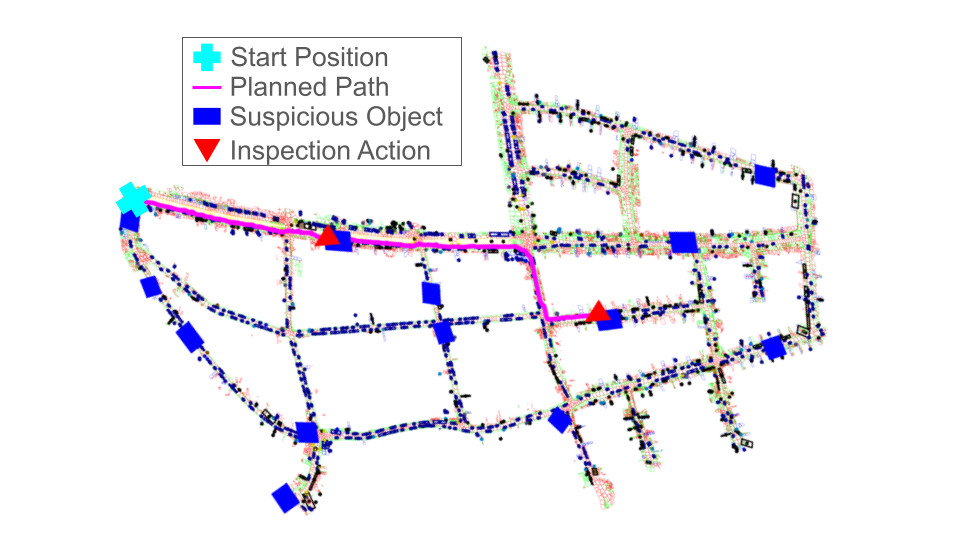}
    \caption{An example plan from the KITTI environment. The robot begins in the top left, and is tasked with inspecting one object (denoted by the red triangle at the end of the trajectory). Along the way, there are numerous objects potentially blocking the path, so we must add at least one to its planning domain. After inspecting and neutralizing this object, the robot can reach its goal.}
    \label{fig:kitti}
\end{figure*}

\noindent\textbf{Object Obstruction:} Next, we investigate the performance of the incremental object solver algorithm in task instances where objects not directly listed in the goal must be inspected in order to solve the task. In this experiment, we give a robot one of two goal types in a scene graph built from the KITTI dataset (Figs. \ref{fig:front} and \ref{fig:kitti}): either \texttt{(Visited $P_i$)} or \texttt{(Safe $O_j$)}. Given the size of the map, satisfying these goals may require the agent to traverse a large distance, but more importantly, if there are obstructing objects in the way, it may be forced to inspect and neutralize them to find a safe path to the goal. As a baseline, we sample 20 goals in the map shown in Fig.~\ref{fig:kitti} using our planner without any of the objects being labeled as \texttt{suspicious}. In this case, they do not obstruct the agent's path, and we find plans in 19 of 20 trials.

Next, we ``activate'' 13 objects in the scene by labeling them as \texttt{suspicious}. A \texttt{suspicious} object has an inflated radius that is only safe for the robot to enter after it has been inspected and neutralized (in the KITTI scene, this radius is large enough to block an entire road as shown by the blue objects in Fig~\ref{fig:kitti}). For the agent to inspect the object, it has to find a pose that is traversable and within range of the object. Then, by taking the inspect action, the object becomes \texttt{safe}, and can be passed. To highlight the importance of object pruning, we attempt to solve these same tasks without using our incremental feedback approach for object pruning~(Sec.~\ref{sec:object_pruning}). Instead, we add all 13 suspicious objects to the scene directly. Using this encoding, the planner only succeeds in finding a plan in 4 out of 20 trials. Inspecting these solutions further reveals that in all 4 of these successful cases, there was a direct path to the goal without inspecting any objects. This result makes sense, as the odds of sampling the correct object to inspect is low without the benefit of geometric information.

Finally, we test our proposed approach of incrementally adding objects to the planning instance (Sec.~\ref{sec:object_pruning}). Our planner solves 12 of the 20 trials, including 9 cases wherein the agent inspected one or more obstructing objects on the way to its goal. Failure to find plans is caused by PDDLStream not successfully finding sequences of inspection poses on the correct side of obstructing objects when several such objects needs to be inspected to find a plan. These experiments further demonstrate the importance of our proposed approach to sparsifying otherwise dense, long-horizon planning problems.
An example plan, where the agent investigates two objects on the way to its goal is shown in Fig.~\ref{fig:kitti}.

\noindent\textbf{Real-World Manipulation:} Finally, we demonstrate our planner in a real-world setting (Fig.~\ref{fig:manipulation-experiment}), using a Boston Dynamics Spot quadruped to build a 3D scene graph in real-time in a university building. In order to demonstrate that our approach is effective on domains different from ``Inspect'', we implement a ``Retrieval" domain, which adds additional \texttt{Pick} and \texttt{Place} actions to the Inspection domain, enabling the robot to move objects around the environment. The \texttt{VisitedPlace} predicate is evaluated from the robot's body position, so that the manipulator's swept volume does not need to be considered when reasoning about places to avoid.
The goal specifies which place an object should be in (e.g., \texttt{(ObjectAtPlace O1 P3)}, denoting a goal state where object \texttt{O1} is in place \texttt{P3}). For each trial we run, we structure the environment such that there is an obstruction preventing the robot from reaching its target object. To solve the task, the robot must move this obstruction out of the way, before retrieving the specified object. We encoded pick, place, inspect, and move skills for the robot.

Like the KITTI domain, we rely on our incremental object adding approach to only add the obstructing object to our high-level planner. Scattered throughout the environment are a number of different objects (Fig.~\ref{fig:manipulation-experiment}), which would lead to an intractable problem if we were to consider them all. We ran ten trials, each time finding plans, and highlight several successful executions in the video supplement\footnote{\url{https://youtu.be/v8fkwLjBn58}}. While the planner reliably finds feasible plans, execution often requires several attempts due to failures when executing the \texttt{Pick} skill and Spot's local planner failing in certain constrained passages.

\section{Related Work}

There has been substantial recent work enabling construction of information-rich 3D scene graphs, initially introduced by~\citet{armeni20193d}. Following works have focused on constructing 3D scene graphs from real-world sensor data~\cite{rosinol2021ijrr}, real-time performance~\cite{Hughes24ijrr-hydraFoundations,bavle2023s}, and improving the higher-level abstractions~\cite{wu2021scenegraphfusion,strader2023indoor}. The strong performance of foundation models on open-vocabulary tasks has led to a series of works on combining open-vocabulary language embeddings with 3D scene graphs~\cite{gu2023conceptgraphs,maggio2024clio,werby2024hierarchical}. These open-vocabulary works all feature object navigation or retrieval tasks executed on real robots, although the task structure is simple and the focus is on mapping natural language to an object grounded in the scene graph.

There has also been recent work focused on applying structured planning domains to 3D scene graphs. \citet{Agia22corl-Taskography} derive a \gls{pddl} representation for task planning from scene graphs, with a focus on using the hierarchical nature of scene graphs to sparsify the representation in order to make planning tractable. Their approach is only guaranteed to produce valid solutions for very specific planning domains, where only constraints between symbols with a clear ``ancestor" relationships are expressible, and it is unclear how to extend this to a more general set of planning tasks. \citet{dai2023optimal} present a method for grounding natural language commands in LTL formula, leveraging the hierarchy of the scene graphs to accelerate planning. Unfortunately the scene graph abstraction may not satisfy the property of downward-refinement~\cite{bacchus1991downward}, breaking many of the assumptions in symbolic task planning. The existence of low-level geometric constraints requires going beyond task planning approaches, and into the realm of \gls{tamp}.

There has been additional work in pruning superfluous elements of scenes to accelerate \gls{tamp}. \citet{silver2021planning} learn to predict which symbols are relevant to a particular \gls{tamp} problem.
\citet{khodeir2023policy} and \citet{vu2024coast} both address PDDLStream's poor scaling as the number of objects grows, with \cite{khodeir2023policy} proposing an algorithm that guides the search through task skeletons based on failures by the motion planner and \cite{vu2024coast} introducing a more intelligent method for instantiating streams. Meanwhile, \cite{learning-to-search} and \cite{iser}
learn to guide search through predictions of relevance or feasibility. These approaches are complimentary to our own, though learning in large scene graphs suffers from issues of generalization.
Outside of learning based approaches, \citet{failure} attempt to guide \gls{tamp} from failed motion plans (much like we do in Sec~\ref{sec:object_pruning}) by adding additional goal conditions, an approach that struggles with large initial object sets.

\section{CONCLUSIONS}

In this work we proposed an approach for enabling and accelerating \gls{tamp} in large scene graphs. We defined characteristics of planning domains that permit the pruning of certain symbols. We then proposed a method for deriving a domain from a Hydra scene graph which has these characteristics, and demonstrate how we prune Places and Objects from the domain. We also proved how the plans we produce from this pruned scene graph are valid and conform to the constraints of the full planning domain. Finally, we demonstrated experimentally how our approach scales with scene graph size, goal complexity, and geometric constraints in several environments, including a scene graph built from the KITTI dataset and real-world execution on a Spot quadraped.

In future work, we hope to demonstrate under what conditions we can extend our pruning method to other domains derived from large-scale scene graphs. Furthermore, augmenting our approach with learned methods for object pruning is a natural extension. The metric-semantic information in the scene graph is potentially a strong signal for a learner to identify further irrelevant symbols in a planning domain.

\renewcommand{\baselinestretch}{0.95}
\begin{center}
    \Large{\textbf{Bibliography}}
\end{center}
\begingroup
\def\bibfont{\footnotesize}
\renewcommand{\clearpage}{}
\renewcommand{\chapter}[2]{}\bibliographystyle{plainnat_nourl}

\begin{thebibliography}{30}
\providecommand{\natexlab}[1]{#1}
\providecommand{\url}[1]{\texttt{#1}}
\expandafter\ifx\csname urlstyle\endcsname\relax
  \providecommand{\doi}[1]{doi: #1}\else
  \providecommand{\doi}{doi: \begingroup \urlstyle{rm}\Url}\fi

\bibitem[Agia et~al.(2022)Agia, Jatavallabhula, Khodeir, Miksik, Vineet,
  Mukadam, Paull, and Shkurti]{Agia22corl-Taskography}
Christopher Agia, Krishna~Murthy Jatavallabhula, Mohamed Khodeir, Ondrej
  Miksik, Vibhav Vineet, Mustafa Mukadam, Liam Paull, and Florian Shkurti.
\newblock Taskography: {{Evaluating}} robot task planning over large {3D} scene
  graphs.
\newblock In \emph{Conf. on Robot Learning (CoRL)}, pages 46--58. {PMLR},
  January 2022.

\bibitem[Armeni et~al.(2019)Armeni, He, Gwak, Zamir, Fischer, Malik, and
  Savarese]{armeni20193d}
Iro Armeni, Zhi-Yang He, JunYoung Gwak, Amir~R Zamir, Martin Fischer, Jitendra
  Malik, and Silvio Savarese.
\newblock {3D} scene graph: A structure for unified semantics, {3D} space, and
  camera.
\newblock In \emph{Intl. Conf. on Computer Vision (ICCV)}, pages 5664--5673,
  2019.

\bibitem[Bacchus and Yang(1991)]{bacchus1991downward}
Fahiem Bacchus and Qiang Yang.
\newblock The downward refinement property.
\newblock In \emph{IJCAI}, pages 286--293, 1991.

\bibitem[Bavle et~al.(2023)Bavle, Sanchez-Lopez, Shaheer, Civera, and
  Voos]{bavle2023s}
Hriday Bavle, Jose~Luis Sanchez-Lopez, Muhammad Shaheer, Javier Civera, and
  Holger Voos.
\newblock S-graphs+: Real-time localization and mapping leveraging hierarchical
  representations.
\newblock \emph{IEEE Robotics and Automation Letters}, 8\penalty0 (8):\penalty0
  4927--4934, 2023.

\bibitem[Blum and Furst(1997)]{blum1997fast}
Avrim~L Blum and Merrick~L Furst.
\newblock Fast planning through planning graph analysis.
\newblock \emph{Artificial intelligence}, 90\penalty0 (1-2):\penalty0 281--300,
  1997.

\bibitem[Bradley and Roy(2024)]{iser}
Christopher Bradley and Nicholas Roy.
\newblock Learning feasibility and cost to guide tamp.
\newblock In \emph{Experimental Robotics}, pages 203--216. Springer Nature
  Switzerland, 2024.

\bibitem[Dai et~al.(2024)Dai, Asgharivaskasi, Duong, Lin, Tzes, Pappas, and
  Atanasov]{dai2023optimal}
Zhirui Dai, Arash Asgharivaskasi, Thai Duong, Shusen Lin, Maria-Elizabeth Tzes,
  George Pappas, and Nikolay Atanasov.
\newblock Optimal scene graph planning with large language model guidance.
\newblock In \emph{IEEE Intl. Conf. on Rob. and Autom. (ICRA)}, pages
  14062--14069, 2024.

\bibitem[Fishman et~al.(2023)Fishman, Kumar, Allen, Danas, Littman, Tellex, and
  Konidaris]{fishman2023task}
Michael Fishman, Nishanth Kumar, Cameron Allen, Natasha Danas, Michael Littman,
  Stefanie Tellex, and George Konidaris.
\newblock Task scoping: Generating task-specific simplifications of open-scope
  planning problems.
\newblock In \emph{PRL Workshop Series \textendash Bridging the Gap Between AI
  Planning and Reinforcement Learning}, 2023.

\bibitem[Garrett et~al.(2020)Garrett, Lozano-P{\'e}rez, and
  Kaelbling]{pddlstream}
Caelan~Reed Garrett, Tom{\'a}s Lozano-P{\'e}rez, and Leslie~Pack Kaelbling.
\newblock {PDDL}stream: Integrating symbolic planners and blackbox samplers via
  optimistic adaptive planning.
\newblock In \emph{Proceedings of the International Conference on Automated
  Planning and Scheduling}, 2020.

\bibitem[Garrett et~al.(2021)Garrett, Chitnis, Holladay, Kim, Silver,
  Kaelbling, and Lozano-P{\'e}rez]{tamp-survey}
Caelan~Reed Garrett, Rohan Chitnis, Rachel Holladay, Beomjoon Kim, Tom Silver,
  Leslie~Pack Kaelbling, and Tom{\'a}s Lozano-P{\'e}rez.
\newblock Integrated task and motion planning.
\newblock \emph{Annual review of control, robotics, and autonomous systems},
  2021.

\bibitem[Ghallab et~al.(2016)Ghallab, Nau, and Traverso]{ghallab2016automated}
Malik Ghallab, Dana Nau, and Paolo Traverso.
\newblock \emph{Automated planning and acting}.
\newblock Cambridge University Press, 2016.

\bibitem[Gu et~al.(2024)Gu, Kuwajerwala, Morin, Jatavallabhula, Sen, Agarwal,
  Rivera, Paul, Ellis, Chellappa, Gan, de~Melo, Tenenbaum, Torralba, Shkurti,
  and Paull]{gu2023conceptgraphs}
Qiao Gu, Ali Kuwajerwala, Sacha Morin, Krishna~Murthy Jatavallabhula, Bipasha
  Sen, Aditya Agarwal, Corban Rivera, William Paul, Kirsty Ellis, Rama
  Chellappa, Chuang Gan, Celso~Miguel de~Melo, Joshua~B. Tenenbaum, Antonio
  Torralba, Florian Shkurti, and Liam Paull.
\newblock Conceptgraphs: Open-vocabulary {3D} scene graphs for perception and
  planning.
\newblock In \emph{IEEE Intl. Conf. on Rob. and Autom. (ICRA)}, pages
  5021--5028, 2024.

\bibitem[Helmert(2006)]{fast_downward}
Malte Helmert.
\newblock The fast downward planning system.
\newblock \emph{Journal of Artificial Intelligence Research}, 26:\penalty0
  191--246, 2006.

\bibitem[Hoffmann(2001)]{fast_forward}
J{\"o}rg Hoffmann.
\newblock Ff: The fast-forward planning system.
\newblock \emph{AI magazine}, 22\penalty0 (3):\penalty0 57--57, 2001.

\bibitem[Hughes et~al.(2024)Hughes, Chang, Hu, Talak, Abdulhai, Strader, and
  Carlone]{Hughes24ijrr-hydraFoundations}
Nathan Hughes, Yun Chang, Siyi Hu, Rajat Talak, Rumaisa Abdulhai, Jared
  Strader, and Luca Carlone.
\newblock Foundations of spatial perception for robotics: Hierarchical
  representations and real-time systems.
\newblock \emph{Intl. J. of Robotics Research}, 2024.

\bibitem[Karpas and Magazzeni(2020)]{karpas2020automated}
Erez Karpas and Daniele Magazzeni.
\newblock Automated planning for robotics.
\newblock \emph{Annual Review of Control, Robotics, and Autonomous Systems},
  2020.

\bibitem[Khodeir et~al.(2023{\natexlab{a}})Khodeir, Agro, and
  Shkurti]{learning-to-search}
Mohamed Khodeir, Ben Agro, and Florian Shkurti.
\newblock Learning to search in task and motion planning with streams.
\newblock \emph{IEEE Intl. Conf. on Rob. and Autom. (ICRA)}, 8\penalty0
  (4):\penalty0 1983--1990, 2023{\natexlab{a}}.

\bibitem[Khodeir et~al.(2023{\natexlab{b}})Khodeir, Sonwane, Hari, and
  Shkurti]{khodeir2023policy}
Mohamed Khodeir, Atharv Sonwane, Ruthrash Hari, and Florian Shkurti.
\newblock Policy-guided lazy search with feedback for task and motion planning.
\newblock In \emph{IEEE Intl. Conf. on Rob. and Autom. (ICRA)}, pages
  3743--3749. IEEE, 2023{\natexlab{b}}.

\bibitem[LaValle(1998)]{lavalle1998rapidly}
Steven LaValle.
\newblock Rapidly-exploring random trees: A new tool for path planning.
\newblock \emph{Research Report 9811}, 1998.

\bibitem[LaValle(2006)]{lavalle2006planning}
Steven~M LaValle.
\newblock \emph{Planning algorithms}.
\newblock Cambridge university press, 2006.

\bibitem[Maggio et~al.(2024)Maggio, Chang, Hughes, Trang, Griffith, Dougherty,
  Cristofalo, Schmid, and Carlone]{maggio2024clio}
Dominic Maggio, Yun Chang, Nathan Hughes, Matthew Trang, Dan Griffith, Carlyn
  Dougherty, Eric Cristofalo, Lukas Schmid, and Luca Carlone.
\newblock Clio: Real-time task-driven open-set {3D} scene graphs.
\newblock \emph{IEEE Robotics and Automation Letters}, 9\penalty0
  (10):\penalty0 8921--8928, 2024.

\bibitem[Oleynikova et~al.(2018)Oleynikova, Taylor, Siegwart, and
  Nieto]{Oleynikova18iros-topoMap}
Helen Oleynikova, Zachary Taylor, Roland Siegwart, and Juan Nieto.
\newblock Sparse {3D} topological graphs for micro-aerial vehicle planning.
\newblock In \emph{IEEE/RSJ Intl. Conf. on Intelligent Robots and Systems
  (IROS)}, 2018.

\bibitem[Rosinol et~al.(2021)Rosinol, Violette, Abate, Hughes, Chang, Shi,
  Gupta, and Carlone]{rosinol2021ijrr}
Antoni Rosinol, Andrew Violette, Marcus Abate, Nathan Hughes, Yun Chang,
  Jingnan Shi, Arjun Gupta, and Luca Carlone.
\newblock Kimera: From slam to spatial perception with {3D} dynamic scene
  graphs.
\newblock \emph{The International Journal of Robotics Research}, 40\penalty0
  (12-14):\penalty0 1510--1546, 2021.

\bibitem[Silver et~al.(2021)Silver, Chitnis, Curtis, Tenenbaum,
  Lozano-P{\'e}rez, and Kaelbling]{silver2021planning}
Tom Silver, Rohan Chitnis, Aidan Curtis, Joshua~B Tenenbaum, Tom{\'a}s
  Lozano-P{\'e}rez, and Leslie~Pack Kaelbling.
\newblock Planning with learned object importance in large problem instances
  using graph neural networks.
\newblock In \emph{Proc. of the AAAI Conf. on Artificial Intelligence},
  volume~35, pages 11962--11971, 2021.

\bibitem[Srivastava et~al.(2014)Srivastava, Fang, Riano, Chitnis, Russell, and
  Abbeel]{failure}
Siddharth Srivastava, Eugene Fang, Lorenzo Riano, Rohan Chitnis, Stuart
  Russell, and Pieter Abbeel.
\newblock Combined task and motion planning through an extensible
  planner-independent interface layer.
\newblock In \emph{IEEE Intl. Conf. on Rob. and Autom. (ICRA)}, pages 639--646.
  IEEE, 2014.

\bibitem[Strader et~al.(2024)Strader, Hughes, Chen, Speranzon, and
  Carlone]{strader2023indoor}
Jared Strader, Nathan Hughes, William Chen, Alberto Speranzon, and Luca
  Carlone.
\newblock Indoor and outdoor {3D} scene graph generation via language-enabled
  spatial ontologies.
\newblock \emph{IEEE Robotics and Automation Letters}, 9\penalty0 (6):\penalty0
  4886--4893, 2024.

\bibitem[Vega-Brown and Roy(2020)]{vega2020task}
William Vega-Brown and Nicholas Roy.
\newblock Task and motion planning is {PSPACE}-complete.
\newblock In \emph{Proc. of the AAAI Conf. on Artificial Intelligence},
  volume~34, pages 10385--10392, 2020.

\bibitem[Vu et~al.(2024)Vu, Migimatsu, and Bohg]{vu2024coast}
Brandon Vu, Toki Migimatsu, and Jeannette Bohg.
\newblock {COAST}: {CO}nstraints {A}nd {ST}reams for task and motion planning.
\newblock In \emph{2024 IEEE International Conference on Robotics and
  Automation (ICRA)}, pages 14875--14881, 2024.

\bibitem[Werby et~al.(2024)Werby, Huang, B{\"u}chner, Valada, and
  Burgard]{werby2024hierarchical}
Abdelrhman Werby, Chenguang Huang, Martin B{\"u}chner, Abhinav Valada, and
  Wolfram Burgard.
\newblock Hierarchical open-vocabulary {3D} scene graphs for language-grounded
  robot navigation.
\newblock In \emph{First Workshop on Vision-Language Models for Nav. and Manip.
  at ICRA}, 2024.

\bibitem[Wu et~al.(2021)Wu, Wald, Tateno, Navab, and
  Tombari]{wu2021scenegraphfusion}
Shun-Cheng Wu, Johanna Wald, Keisuke Tateno, Nassir Navab, and Federico
  Tombari.
\newblock Scenegraphfusion: Incremental {3D} scene graph prediction from
  {RGB-D} sequences.
\newblock In \emph{Proc. of the IEEE/CVF Conf. on Comp. Vision and Pattern
  Recog.}, pages 7515--7525, 2021.

\end{thebibliography}

\endgroup

\section*{Appendix}
\textbf{Proof of Prop~\ref{prop:ignore_strongly_redundant}.}
Consider $\pi \in \Pi_R$. If $\pi$ does not contain any actions parameterized by $x$, then the same plan $\pi$ is also a valid solution for $R'$. Consider the alternative, where $\pi$ does contain an action parameterized by $x$. By Def.~\ref{def:strong_redundant}, there is another plan $\pi'$ with equivalent motion sequence not parameterized by $x$, which is a valid solution for $R'$.

Now that we have shown that $\Pi_{R'}$ is not empty, we need to show that any valid plan for $R'$ is valid for $R$. Consider plan $\pi = [a_1, ..., a_N] \in \Pi_{R'}$ with corresponding state plan $\mathcal{I}_{\pi} = [\mathcal{I}_0, ..., \mathcal{I}_N]$. If the addition of facts $\mathcal{F}$ parameterized by $x$ make $\pi$ invalid, then there must exist a state $\mathcal{I}_k$ such that $\mathcal{I}_k \cup \mathcal{F} \notin \text{Pre}(a_{k+1})$, which means that $a_{k+1}$ is parameterized by a symbol that did not exist in $R'$. Only a universal or existential quantifier in $\text{Pre}(a_{k+1})$ can cause $a_{k+1}$ to be parameterized by an additional symbol. Adding additional facts cannot turn an existentially quantified formula from true to false. By Definition \ref{def:strong_redundant}, $a_{k+1}$ does not have any universal quantifiers that can be parameterized by $x$ in its precondition. Thus $\pi$ must be valid for $R$.\qed

\noindent\textbf{Proof of Prop~\ref{prop:redundant_places}.}
First note that no actions in this domain have universal quantifiers, so we only need to check Definition \ref{def:strong_redundant}.1 to show that a symbol is redundant. Consider a place $p$ such that \texttt{(not (VisitedPlace $p$))} appears in the CNF of the goal. If $p$ parameterizes \texttt{moveRelaxed}, then \texttt{(VisitedPlace $p$)} is in the effects, violating the goal. Since  \texttt{moveRelaxed} is the only action that can be parameterized by a place, no plan can parameterize $p$ and Definition \ref{def:strong_redundant}.1 is trivially satisfied.

Next, consider a place $p$ that does not parameterize any initial or goal facts. For any plan $\pi$ with an action parameterized by $p$, let $a_k^{\mathcal{P}}$ denote an action parameterized by places $\mathcal{P}$, including $p$. Plan $\pi'$ where $a_{k}^\mathcal{P}$ is replaced by $a_{k}^{\mathcal{P}\setminus p}$ is also valid, since state plans $\mathcal{I}_{\pi}$ and $\mathcal{I}_{\pi'}$ only differ by a \texttt{(VisitedPlace $p$)} fact, and no action preconditions or goals involve this fact.
As a result the command sub-sequence corresponding to $a_{k}^\mathcal{P}$ is also valid for $a_{k}^{\mathcal{P}\setminus p}$. So, for any $\pi$ parameterized by $p$, we can construct $\pi'$ that has an equivalent motion sequence but does not parameterize $p$; thus $p$ is redundant. \qed

\end{document}